# MULTI-VIEW STEREO WITH SEMANTIC PRIORS

E.-K. Stathopoulou, F. Remondino

3D Optical Metrology (3DOM) unit, Bruno Kessler Foundation (FBK), Trento, Italy
Web: http://3dom.fbk.eu
Email: <estathopoulou><remondino>@fbk.eu

**Commission II, WG II/8**

**ABSTRACT:**

Patch-based stereo is nowadays a commonly used image-based technique for dense 3D reconstruction in large scale multi-view applications. The typical steps of such a pipeline can be summarized in stereo pair selection, depth map computation, depth map refinement and, finally, fusion in order to generate a complete and accurate representation of the scene in 3D. In this study, we aim to support the standard dense 3D reconstruction of scenes as implemented in the open source library OpenMVS by using semantic priors. To this end, during the depth map fusion step, along with the depth consistency check between depth maps of neighbouring views referring to the same part of the 3D scene, we impose extra semantic constraints in order to remove possible errors and selectively obtain segmented point clouds per label, boosting automation towards this direction. In order to reassure semantic coherence between neighbouring views, additional semantic criterions can be considered, aiming to eliminate mismatches of pixels belonging in different classes.

## 1. INTRODUCTION

Obtaining precise 3D information from images with photogrammetric and computer vision techniques has become a common practice in applications such as city modelling, structure monitoring, indoor navigation or heritage documentation, often preferred over costly laser scanning solutions. The tremendous increase in the last decades of the computational power along with the new released sensor technologies have facilitated the recent advances in all main steps of the 3D reconstruction workflow (Agarwal et al., 2009; Lourakis and Argyros, 2009; Wu et al., 2011; Rothermel et al., 2012; Schoenberger and Frahm, 2016; Remondino et al., 2017) as well as the implementation of these methods to various challenging case studies (Stathopoulou et al., 2015; Menna et al., 2016; Remondino et al., 2016).

The typical 3D reconstruction pipeline can generally be divided into two main parts: image orientation (so-called Structure from Motion – SfM) (Ozyesil et al., 2017) and dense image matching (often called Multi-View Stereo – MVS) (Remondino et al., 2014; Furukawa and Hernandez, 2015). SfM refers to the camera pose estimation and sparse point cloud generation based on the accurate detection and matching of homologous image features.

On the other hand, MVS algorithms are addressing the last part of the photogrammetric chain-flow aiming to generate a densified point cloud by pairwise or multi-view matching of every pixel of the images (disparity or depth calculation) and successively triangulate in the 3D space. Dense 3D reconstruction from images is an active research area in the latest decades. Great efforts have been made and algorithms have reached maturity in terms of efficiency, scalability and accuracy. Semi-Global Matching (Hirschmüller, 2005) and patch-based methods (Barnes et al., 2009; Bleyer et al., 2011) are the most popular among them. However, dense image matching methods may still produce many errors, noisy clouds or even fail in uncontrolled environments' cases, due textureless or reflective surfaces, commonly present in terrestrial scenarios (e.g. urban, indoor, cultural heritage).

Meanwhile, recent trends in computer vision and data science have led to an extensive usage of machine and deep learning methods on images and 3D point clouds for classification, scene semantic segmentation or object detection in applications such as robotics, geospatial or cultural heritage (Poux et al., 2017; Weinmann et al., 2017; Grilli and Remondino, 2019). The term semantic segmentation refers to the assignment of a predicted label for each single image pixel in a semantic meaningful way. Thus, semantic segmentation research aims towards full scene understanding using object knowledge.

### 1.1 Aim of the paper

By leveraging 3D reconstruction and semantic segmentation, the underlying semantic information can potentially constrain the shape of the scene objects, assuming disparity smoothness within the same class and facilitate thus the depth calculation in cases where more than one possible depth value could be assigned.

Based on the work of Stathopoulou and Remondino (2019) on enhancing the various steps of the photogrammetric pipeline with previously obtained semantic information (Figure 1), we hereby focus on the depth/3D dense point cloud computation step. Hence, we aim to enhance the MVS procedure with semantic information in order to improve the quality of the achieved 3D results or derive (selected) semantically segmented 3D point clouds.

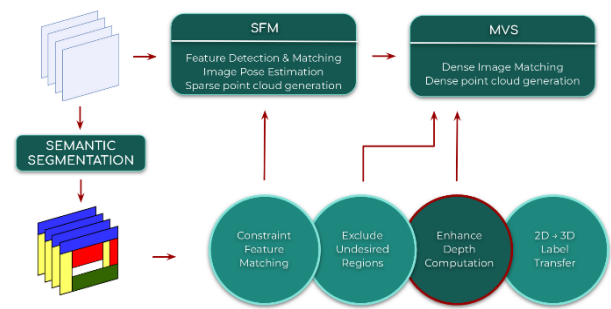

Figure 1: The proposed semantic photogrammetry reconstruction pipeline where semantic priors are incorporated to support the 3D results (modified from Stathopoulou and Remondino, 2019).

For instance, pixel pairs with labelling inconsistency should be considered as high cost matches or pixels belonging to undesired or poorly defined and fuzzy parts of the scene e.g. sky, obstacles, trees will be penalized differently than the salient ones, or even being excluded, improving this way the robustness of the reconstruction. More particularly, in this article, we jointly infer the depth information and the semantic smoothness criterion during the depth map fusion step, generating selective point clouds based on their semantic label. Along with this, we also consider imposing semantic consistency constraints between neighbouring views in order to avoid potential mismatches across different classes.

These semantic priors are provided through the same methodology used by Stathopoulou and Remondino (2019), i.e. by training a Convolutional Neural Network (CNN) (Jégou et al., 2017) on images of historic building façades.

In the next sections, related works on dense reconstruction and semantic depth is discussed (Section 2), followed by the description of the MVS algorithm that our application is based on (Section 3). Details of our implementation and experiments are given in Section 4, while the final conclusions and future work are outlined in Section 5.

## 2. RELATED WORKS

In photogrammetry, depth calculation and 3D dense point cloud generation are usually performed pairwise using stereo matching techniques, usually divided into local and global ones. Semi Global Matching (SGM) (Hirschmüller 2005 and 2008) was the breakthroughs global method and got popularity due to its efficiency particularly in aerial and industrial applications. Several variations have been built upon SGM (Rothermel et al, 2012; Spangenberg et al., 2014; Sinha et al., 2014; Scharstein et al., 2017; Roth and Mayer, 2019) implying geometric or other constraints or even using CNNs (Seki and Pollefeys, 2016). SGM approximates the minimization of a 2D Markov Random Field (MRF) energy function by performing cost aggregation along various 1D paths for each pixel, making use also of depth smoothness constraints, i.e. making the assumption that neighbouring pixels are very likely to have the same depth. Thus, SGM performs well for textured scenes and aerial cases, yet it often fails on slanted surfaces and wide-baseline configurations commonly used in terrestrial applications (Roth and Mayer, 2019). This is due to the so-called fronto-parallel bias i.e. the assumption that the image plane and object plane are parallel, or, in other words that the depth is almost constant within the searching window.

On the other hand, patch-based MVS methods are based on depth propagation between neighbouring pixels. It was initially introduced by Bleyer et al. (2011), based on the concept of Barnes et al. (2009), in order to solve the stereo matching problem and reduce the effect of the fronto-parallel bias (Roth and Mayer, 2019). However, nowadays it is commonly used in large-scale multi-view applications because of its efficiency and scalability (Shen, 2013; Zheng et al, 2014; Galliani et al., 2015; Schönberger et al., 2016). Furukawa and Ponce (2009) introduced a revolutionary patch-based algorithm for MVS reconstruction (PMVS): starting from a sparse set of matched keypoints, patches are initialized and repeatedly expanded based on visibility constraints in order to reconstruct the surface around them. Scalability problem of PMVS was tackled by image clustering (Furukawa et al., 2010). Generally, patch-based stereo methods initialize each pixel with a random disparity and a randomly slanted plane and iteratively propagates them to neighbouring pixels. While one of the core challenges in patch-based MVS is the view selection, commonly formed as a probabilistic model (Zheng et al., 2014), depth calculation as well as filtering and fusion are still to be optimized in order to achieve highly accurate depth maps. A recent benchmark to evaluate multi-view stereo methods was presented in Schöps et al. (2017).

Image understanding, classification and segmentation has become a very fruitful research topic for several applications ranging from autonomous driving to cultural heritage applications (Finman et al., 2014; Martinovic et al., 2015; Zhang et al., 2015). Semantic labelling is nowadays performed using machine/deep learning techniques, with Fully Convolutional Neural Networks (FCNs) being a pioneering work, proving great performance in 2D segmentation and outperforming other methods (Long et al., 2015; He et al., 2017).

The combination of dense 3D reconstruction algorithms with semantic labels can be exploited in order to produce precise 3D scene representation (Schneider et al., 2016). Whereas several methods impose geometric priors (Gallup et al., 2010; Scharstein et al., 2017) during the depth map calculation, up to our knowledge few works exist in the literature integrating the semantic context into dense depth estimation algorithms. In most of these scenarios, semantics usually imply common sharing of geometric properties and are introduced as object knowledge information constraints (Guney and Geiger, 2015). For instance, the assumption that pixels belonging to the same class must necessarily share the same disparity value is made to guide depth computation for challenging surfaces (Chen et al., 2014). Although same class labels can potentially imply same colour or shape properties (i.e. conditioning on the semantic class of the object, category-based shape information is imposed), this assumption may however do not hold in all cases and therefore generate misleading results.

In Stathopoulou and Remondino (2019) the concept of integrating the semantic information to boost the entire photogrammetric 3D reconstruction pipeline was introduced, focusing on sparse feature matching, automatic mask generation and label transferring to 3D point cloud (Figure 2). Based on this work, in this paper we focus on the dense reconstruction / MVS part, proposing a semantic stereo approach that integrates semantic scene priors to a patch-based stereo algorithm. In particular, we employ the OpenMVS[1] (Open multi-view stereo reconstruction) library, including our semantic constraints during merging step of the computed depth maps.

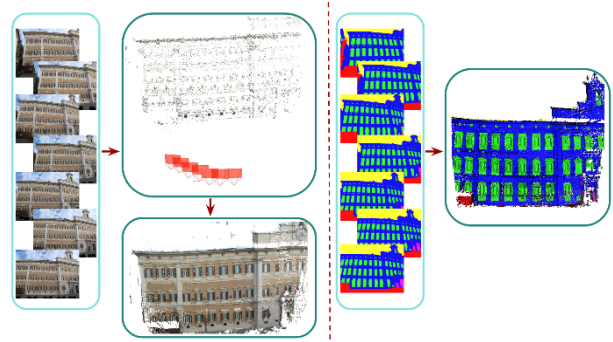

Figure 2: Input set of images and the network orientation with the resulting sparse point cloud. Image labels are projected to 3D, producing a semantically segmented dense point cloud, as obtained by the pipeline presented in Stathopoulou and Remondino (2019).

## 3. PATCH-BASED MVS

A patch-based algorithm generally calculates the depth $d$ for each scene pixel $p$ by repetitively applying spatial propagation starting from random initialization.

---

[1] https://github.com/cdcseacave/openMVS

OpenMVS enfolds a dense reconstruction module based on patch-based stereo, propagation and random refinement (Barnes et al., 2009; Shen, 2013). Following Shen (2013), the main steps for dense depth calculation can be summarized as:

1) *Stereo pair selection*: views are chosen based on intersection angles and visibility criteria. This is a crucial step and it should be carefully designed, especially in the case of unordered images. As stated in Shen (2013), for every image, a good potential pair should fulfil the dual criterion of (i) similar viewing direction and (ii) adequate baseline length. The best angles between the principal viewing directions of two cameras are selected using the visibility of the already available sparse 3D points delivered from the image orientation procedure. An acceptable principal view direction angle $\theta$ is between 5° and 60°. For the images that fulfil this criterion, the median distance $\bar{d}$ between neighbouring optical centres is computed and acceptable distances are considered to be the ones whose $d > 2\bar{d}$ or $d < 0.05\bar{d}$.

2) *Depth map computation*: for every eligible stereo pair, an initial depth map is approximated by interpolating the 3D sparse point cloud of the image orientation procedure. The depth map is then computed using randomly assigned slanted support planes to each pixel $p$ (Bleyer et al., 2011). A support plane is defined as a tangent plane of the local scene surface, represented by a 3D point $X$ and its normal $n$ (Figure 3).

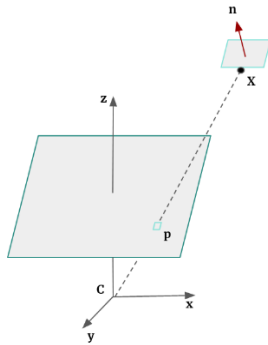

Figure 3: Support plane of an image pixel $p$, represented as a 3D point $X$ and its normal $n$ (adapted from Shen, 2013).

The point $X$ lies on the viewing ray of $p$. Given the camera intrinsic parameters $K$, for any randomly selected depth value $\lambda$:

$$X = \lambda K^{-1} p$$

along with a randomly assigned plane normal $n$.

In case of high-resolution images, this random initialization is likely to have at least one good guess for each depth value. Since the homography mapping between images is already known form the image orientation phase, potential pixel correspondences are established. For initialization, the aggregated matching cost is calculated using the Zero mean Normalized Cross Correlation (ZNCC) measure, which integrates the subtraction of the local mean to the NCC and tends thus to provide more robustness. During each iteration, two operations are being performed on each image pixel:

- spatial propagation: it compares assigned planes between neighbouring pixels in order to ensure depth smoothness among them;
- random assignment: it tries to further reduce the initial calculation of the matching cost by tuning the parameters of the various randomly assigned planes (Bleyer et al.,

2011). In such a way, pixels with high aggregated matching costs are removed.

3) *Depth map refinement (filtering)*: consistency between multiple views is enforced for every map in order to refine the depth values, remove errors and reassure consistency among neighbouring views referring to the same area of the scene. To this end, each point $X$ is reconstructed in 3D using its depth value $\lambda$, the camera intrinsic parameters $K$, the rotation matrix $R$ and the camera centre $C$:

$$X = \lambda R^T K^{-1} p + C$$

Then, it is back projected to the neighbouring 2D views and it is kept only if its depth is coherent over enough neighbouring images. In other words, if the depth differences are close enough for sufficient number $k$ of images, then the point is considered as a valid scene point, otherwise it is discarded.

4) *Depth map merging (fusion)*: the various depth maps that view the same part of the scene are (i) fused together to remove the redundant depth values for every back projected 3D point and (ii) projected to the 3D space in order to create a smooth and unique dense cloud. While fusing, the so-called neighbouring depth map test is performed. Again, pixels are projected to the 3D space and back projected to the neighbouring views, merging just the depth values that are considered to be close enough, in the same fashion like in the previous depth filtering step. The remaining valid depth values are subsequently projected to 3D delivering a single fused point cloud.

## 4. SEMANTIC PATCH-BASED MVS

### 4.1 Dataset preparation

For our experiments, we use images and their corresponding labelled data as introduced in Stathopoulou and Remondino, (2019). The dataset includes several pictures of cultural heritage building facades across various Italian cities (Figure 4). Data labelling has been performed manually in order to generate precise ground truth data for the training of a CNN network for automatic semantic segmentation[2], achieving satisfying scores. Multiple views of the same building facades are available, enabling robust 3D reconstruction of the scenes. The labelled images include five classes sharing certain similarities, namely "building", "sky", "obstacle", "window" and "door".

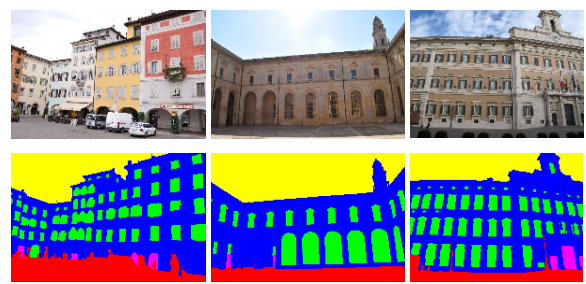

Figure 4: Example images (above) and their labelled equivalent (below) of out labelled dataset. The classes correspond to the following colours: "sky"=yellow, "building" = blue, "window"=green and "obstacle"=red.

### 4.2 Implementation details

As mentioned before, the goal of the work is to couple the dense reconstruction step of the photogrammetric pipeline with priors provided by image semantic classification. Therefore, the aim is twofold: (i) to perform a per-class 3D reconstruction and improve 3D results using semantic guidance; (ii) to eliminate potential mismatches across pixels that belong to diverse semantic classes, reducing this way the possible gross errors.

---

[2] https://github.com/GeorgeSeif/Semantic-Segmentation-Suite

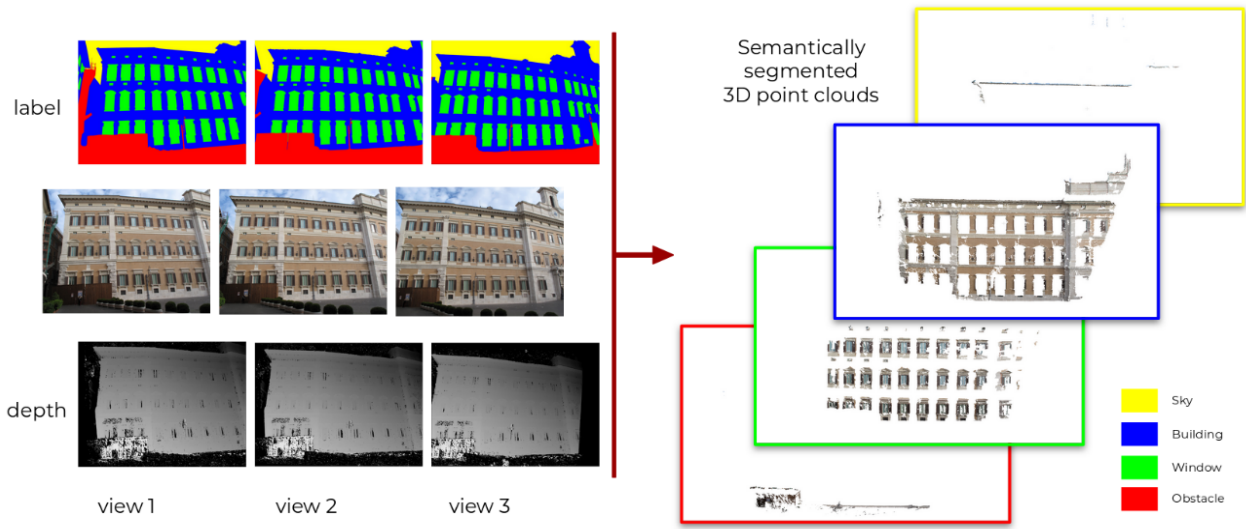

Figure 5: Original input images with their accompanying labels and the calculated depth maps (left). During depth map fusion and point projection, semantic constraints are applied in order to reconstruct just the labels of interest. In this example, one point cloud for each present label is generated and shown (right).

OpenMVS library takes as input the camera orientation parameters along with the sparse 3D point cloud as estimated in most common open-source image orientation / SfM tools (e.g. OpenMVG[3], COLMAP[4], VisualSfM[5], etc.). Although OpenMVS can deliver as final product a refined and textured 3D mesh, in this study we focus and adapt its dense 3D point cloud generation method (Shen, 2013).

Our approach, as starting point, assumes an accurate image orientation (interior and exterior) and 3D sparse point cloud, followed by the generation of undistorted images. Subsequently, our adjusted MVS procedure takes corresponding labelled data and their link to the original and undistorted images with a direct pixel to pixel mapping.

Using the pixel level semantic labels, our semantic consistency constraint is asserted during the last step of the algorithm, i.e. the depth map fusion. Thus, along with depth consistency check between depth maps of neighbouring views referring to the same part of the 3D scene (Section 3), an extra semantic check is performed. In this a way, the included semantic constraint (i) removes possible mismatches over pixels that belong to different semantic classes and (ii) facilitates the generation of separate 3D dense point clouds for each semantic class by reconstructing only the depth of the pixels that are assigned to a certain label. Figure 5 depicts the latter concept: a set of images with known orientation parameter and labelled information is imported in OpenMVS and the depth maps are computed and filtered as explained above. The corresponding labels are linked to the original images and a respective 3D point cloud for each class is generated according to the semantic criterions set each time. Hence, undesired regions (such as fuzzy, sparse or reflective surfaces) of the scene can be automatically excluded (Figure 6). This method, outlined in Figure 7, follows the same line of though with other masking techniques, yet is applied in a semantically meaningful way, providing automation and robustness in the resulting segmented point clouds.

Furthermore, semantic consistency constraints over images can also be applied, by reconstructing in 3D only the depth of the corresponding pixels that have the same label value within the paired views. Using this restriction, pixels with inconsistent labels are excluded from the reconstruction, eliminating possible gross error mismatches across the classes. The resulting point clouds will contain eventually a smaller set of points, thought they have the potential to be more robust.

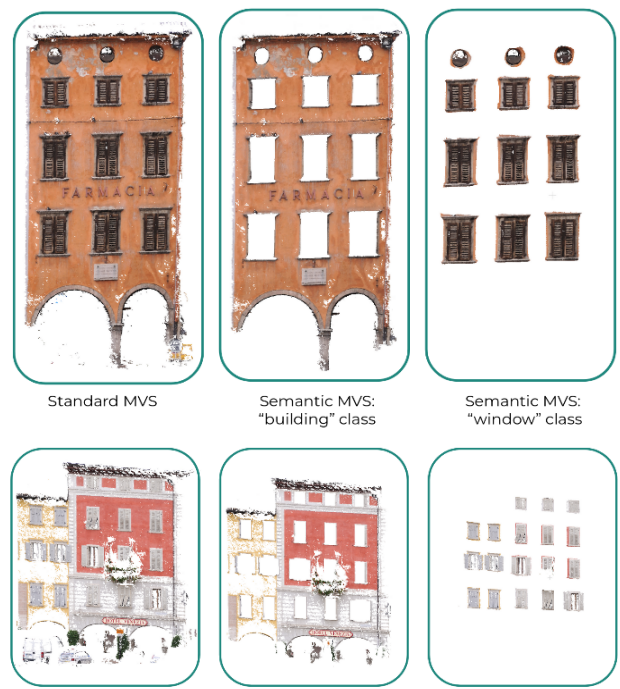

Figure 6: Dense point cloud of a building façade using the standard patch-based MVS algorithm as implemented by OpenMVS (left) and after applying the proposed semantic constraints keeping only points under "building" (middle) and "window" (right) classes. Noisy and undesired 3D points under label "obstacle" or "sky" are automatically excluded.

---

[3] https://github.com/openMVG/openMVG
[4] https://github.com/colmap/colmap
[5] http://ccwu.me/vsfm/

```
Algorithm 1: Semantic patch-based MVS
Input  : set of images
         corresponding label data
         image orientations
         sparse 3D point cloud
Output: Semantically segmented dense point cloud
stereo pair selection;
depth map computation;
depth map refinement;
while depth map merging do
    if d is similar to k neighbouring views then
        if pixel_label = class_of_interest then
            keep d;
            project point in X 3D;
        else
            discard;
        end
    else
        discard;
    end
end
```

Figure 7: Summary of our approach, adding semantic constrains during the depth map fusion step of patch-based MVS.

## 5. CONCLUSIONS AND FUTURE WORKS

In this paper semantic priors were integrated to the dense reconstruction of a scene, adjusting the open-source OpenMVS library. The dense reconstruction / MVS algorithm is a patch-based multi-view stereo method. The standard dense reconstruction procedure is adapted in order to ingest semantically labelled images and associate them with the original images. Eligible pair selection, depth estimation and filtering are integrated with an additional semantic constraint before the final depth map fusion step.

Results are considered promising, since semantic constraints are providing automation in selecting the class of interest to be reconstructed or to filter out unwanted areas (e.g. windows, sky, etc.) or to generate a semantically classified 3D point cloud. The proposed MVS with semantic consistency checks between neighbouring views can potentially generate point clouds that are more robust, since possible mismatches over pixels that belong to variant classes can be excluded from the reconstruction.

Future works include the application of semantic consistency checks also during the depth computation and filtering steps, by penalizing label variations in order to obtain high quality and error free depth maps before the final fusion step.